# DAFOS: 동적 적응형 팬아웃 최적화 샘플러

## DAFOS: Dynamic Adaptive Fanout Optimization Sampler

Irfan Ullah*, Young-Koo Lee †

*Department of Computer Science and Engineering, Kyung Hee University (Global Campus), Republic of Korea

†College of Software, Kyung Hee University (Global Campus), Republic of Korea

{irfan, yklee}@khu.ac.kr

‡Corresponding author: Young-Koo Lee (yklee@khu.ac.kr)

### Abstract

Graph Neural Networks (GNNs) are becoming an essential tool for learning from graph-structured data, however uniform neighbor sampling and static fanout settings frequently limit GNNs' scalability and efficiency. In this paper, we propose the Dynamic Adaptive Fanout Optimization Sampler (DAFOS), a novel approach that dynamically adjusts the fanout based on model performance and prioritizes important nodes during training. Our approach leverages node scoring based on node degree to focus computational resources on structurally important nodes, incrementing the fanout as the model training progresses. DAFOS also integrates an early stopping mechanism to halt training when performance gains diminish. Experiments conducted on three benchmark datasets, ogbn-arxiv, Reddit, and ogbn-products, demonstrate that our approach significantly improves training speed and accuracy compared to a state-of-the-art approach. DAFOS achieves a 3.57x speedup on the ogbn-arxiv dataset and a 12.6x speedup on the Reddit dataset while improving the F1 score from 68.5% to 71.21% on ogbn-arxiv and from 73.78% to 76.88% on the ogbn-products dataset, respectively. These results highlight the potential of DAFOS as an efficient and scalable solution for large-scale GNN training. Implementation of DAFOS and SoTA is variable at Github: DAFOS .

## I. Introduction

Recently, graph neural networks (GNNs) have become a prominent tool for processing graph-structured data, such as social networks, biology, and recommendation systems. GNNs enable the aggregation of neighborhood information through multi-layer message passing allowing models to learn both local and global context in the graph [1]–[5]. However, a key challenge in training GNNs is balancing the computational efficiency with the model's capacity to learn useful context and information. The fanout largely influences this challenge—the number of neighbors sampled per node at each layer of the GNN. Selecting a fixed fanout value throughout the training process can lead to inefficiencies. Small fanout sizes may ignore valuable graph structure, while large fanouts can overwhelm computational resources and slow down training.

In this paper, we propose a novel sampler, Dynamic Adaptive Fanout Optimization Sampler (DAFOS) that introduces adaptive fanout adjustment and node scoring to improve GNN training efficiency and convergence speed. The approach dynamically adjusts the fanout size at the end of each epoch by monitoring the average model's loss, ensuring that the GNN gradually expands its receptive field as needed. By starting with a smaller fanout and incrementally increasing it when the loss plateaus, DAFOS avoids unnecessary computational complexity in the early stages of training. Moreover, our approach incorporates node scoring, prioritizing high-degree nodes, accelerating convergence by focusing on structurally important nodes early in training, and optimizing resource use for faster learning and better performance.

Dynamic fanout adjustment and node scoring enable faster convergence by effectively managing the balance between learning efficiency and accuracy. Additionally, DAFOS employs early stopping based on F1 score improvements, ensuring that training halts when performance gains diminish. This comprehensive approach results in faster training times and improved model performance, particularly in large-scale graphs.

## II. Related Work

In recent years, a great deal of research into GNNs has been conducted, with notable advancements in scalability and training efficiency. Early approaches like GraphSAGE [6] introduced a fixed neighborhood sampling strategy to scale GNNs to large graphs. While this approach reduces computational complexity, it relies on static fanout values, which can limit the model's flexibility during training. Models with fixed fanouts may struggle to fully capture the graph's complexity or unnecessarily consume excessive resources.

Subsequent research has focused on more advanced sampling techniques. FastGCN [1], [2] introduced importance sampling, focusing on reducing the computational cost of large-scale graph learning, but it did not incorporate adaptive mechanisms to adjust sampling dynamically. Similarly, LADIES [1], [4] utilized layer-wise sampling to approximate node distributions, improving training efficiency but still lacked a dynamic adjustment strategy based on real-time model performance.

The idea of Node Scoring has gained attention for accelerating GNN training by prioritizing important nodes. For instance, GraphSAINT [5] uses subgraph sampling to improve scalability, but its node prioritization mechanism is not adaptive and does not consider changes in model performance. Our approach extends this idea by incorporating dynamic node scoring, where nodes are prioritized based on their structural importance early in the training process. This ensures that nodes with higher influence in the graph, such as high-degree, are sampled more frequently, accelerating the model's ability to learn important patterns.

Other research works, such as AdaEdge [3], automated certain aspects of GNN training by adapting edge weights based on scale, but did not consider adaptive fanout adjustment during training. Furthermore, Cluster-GCN [7] improved scalability by partitioning the graph into clusters, but the fixed partitioning approach limits its adaptability. In contrast, our approach incorporates dynamic fanout adjustment, incrementally increasing the fanout based on model loss. This allows the model to gradually expand its receptive field as training progresses, ensuring efficient resource utilization. This balances efficiency and performance by adapting to the model's needs. By starting with a smaller fanout, we maintain computational efficiency during early epochs, while expanding the fanout when the loss plateaus helps capture more global information without overwhelming memory. Additionally, the integration of node scoring prioritizes structurally important nodes, accelerating learning by focusing on the most critical nodes early in training.

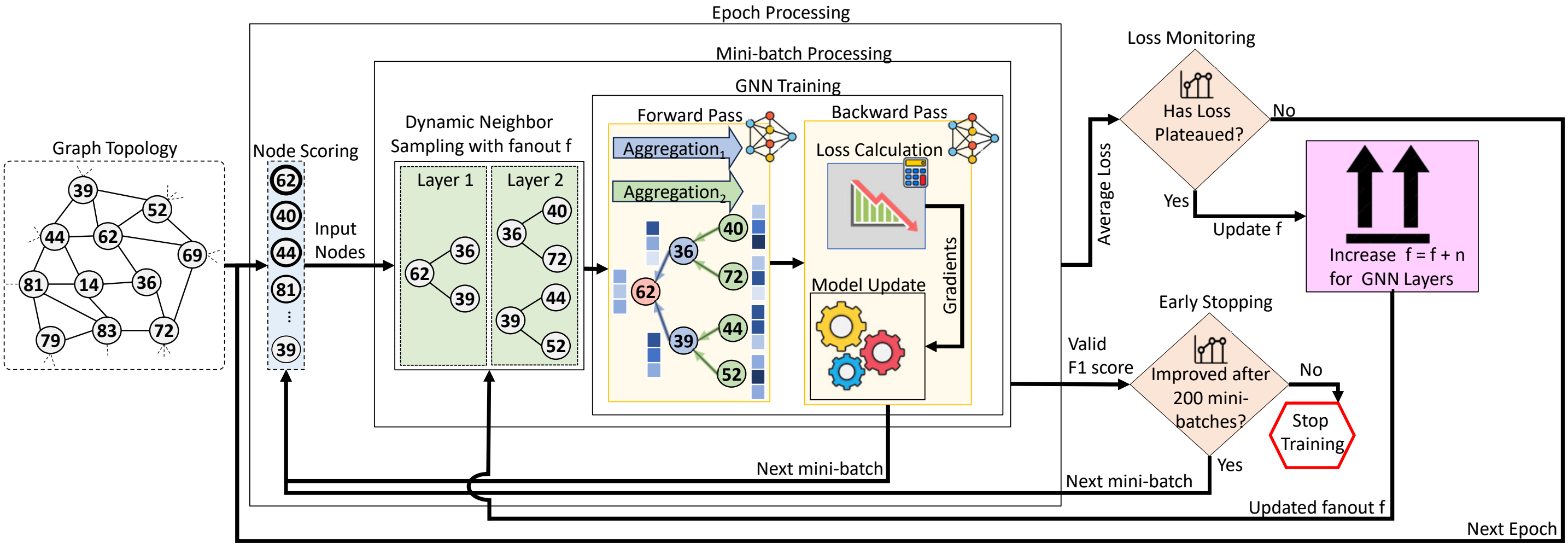

Fig. 1: The DAFOS Architecture begins by selecting a batch of important nodes, followed by forward and backward passes in each GNN layer. At the end of each epoch, the loss is monitored, and if it plateaus, the fanout is increased. The node scoring mechanism prioritizes high-degree nodes during the early stages to accelerate learning.

Dynamic fanout adjustment not only improves computational efficiency but also speeds up model convergence by preventing unnecessary overfitting in early stages and enabling more extensive information aggregation as needed.

## III. Proposed Methodology

This section explains the Dynamic Adaptive Fanout Optimization Sampler (DAFOS), an approach designed to optimize GNN training by dynamically adjusting fanout based on model performance and prioritizing important nodes. The approach is outlined in Figure 1. In DAFOS, node importance is determined solely by node degree, which measures how connected a node is in the graph. The node scoring function is defined as:

$$S(v_i) = d(v_i)$$

Where $S(v_i)$ represents the score of node $v_i$ while $d(v_i)$ represents the degree of node $v_i$. Once node scoring is applied, the GNN proceeds with the message passing across multiple layers, where neighbors are sampled based on the fanout at each layer. The forward propagation in each layer is described as:

$$H^{(l+1)} = \sigma\left(\sum_{v_j \in (\text{sample}(\mathcal{N}(v_i), f^l))} W^{(l)} H_j^{(l)}\right)$$

Where $H^{(l+1)}$ represents node representations at layer $l+1$, $H_j^{(l)}$ represents the node representation of neighbor $v_j$ at layer $l$, $\mathcal{N}(v_i)$ represents the neighbors for node $v_i$, $W^{(l)}$ represents the weight matrix at layer $l$, $\sigma$ is the activation function (e.g., ReLU), and $f^l$ is the fanout for sampling $f^l$ nodes at layer $l$.

By starting with a smaller fanout in the initial epochs, we minimize computational overhead while focusing on local structures. The node scoring mechanism, which prioritizes high-degree nodes, accelerates training by ensuring that the most structurally important nodes are processed first, leveraging their influence on the graph's overall structure.

To improve further training efficiency, DAFOS proposes dynamic fanout adjustment. This idea is based on the concept of progressive model complexity. By gradually expanding the receptive field as the model converges, we enable it to capture global graph structures incrementally. The fanout is incrementally increased at the end of each epoch based on the model's loss. If the change in loss between consecutive epochs is below a predefined threshold, the fanout is adjusted as follows:

$$f^{(t+1)} = \begin{cases} f^{(t)} + \Delta f & \text{if } \left|\mathcal{L}^{(t)} - \mathcal{L}^{(t-1)}\right| < \epsilon \\ f^{(t)} & \text{otherwise} \end{cases}$$

Where $f^{(t+1)}$ represents the fanout for epoch $t+1$, $f^{(t)}$ is the fanout for epoch $t$, $\mathcal{L}^{(t)}$ represents the loss at epoch $t$,$\mathcal{L}^{(t-1)}$ represents the loss at epoch $t-1$, $\epsilon$ is the threshold for determining loss plateau, and $\Delta f$ is the increment in fanout when the loss has plateaued. This dynamic adjustment ensures that the model samples more neighbors and gathers more global information as training progresses. The loss is calculated using the cross-entropy loss as follows:

$$\mathcal{L} = -\frac{1}{N}\sum_{i=1}^{N}\left[y_i \log \hat{y}_i + (1 - y_i)\log(1 - \hat{y}_i)\right]$$

Where $\mathcal{L}$ is the cross-entropy loss, $N$ is the number of training nodes, $y_i$ refers to true label of node $i$, and $\hat{y}_i$ represents the predicted label of node $i$. The average loss is monitored after each epoch to determine whether the fanout should be increased.

Finally, DAFOS includes an early stopping mechanism based on the F1 score. Training halts if the F1 score improvement is below a threshold over a certain number of mini-batches. The early stopping condition is defined as: Stop training if $\left(F1^{(t)} - F1^{(t-n)}\right) < \delta$ for n consecutive mini-batches, where $F1^{(t)}$ is F1 score at mini-batch $t$, $F1^{(t-n)}$ is the F1 score $n$ mini-batches earlier, $\delta$ is the threshold for minimum F1 score improvement, and $n$ is the number of consecutive mini-batches. This ensures that training stops when no significant improvements are seen, optimizing training time and preventing overfitting.

## IV. Results and Discussion

The experiments were conducted on a system running Ubuntu 20.04.6 LTS with 16 AMD Ryzen 7 5800X 8-core processors, 94GB of RAM, and an NVIDIA GeForce RTX 3060 GPU with 12GB of memory, utilizing CUDA 12.4 for computations. The datasets used in this evaluation were ogbn-arxiv[1], Reddit [2], and ogbn-products[3] three benchmark graph datasets with distinct structural properties, and the models were trained and evaluated using the latest Deep Graph Library (DGL) version 2.2.1[4] framework as the state-of-the-art (SoTA) implementation.

In our setup, the fanout values for the two GNN layers were initialized at 10 for layer 1 and 15 for layer 2, with the fanout incremented $(\Delta f)$ by 5 after each epoch when the loss difference between consecutive epochs was less than the threshold $\epsilon = 0.01$. The GNN model utilized hidden layers of size 256 and a learning rate of 0.01. The training was performed with a batch size of 1024, 300 epochs, and each experiment was repeated 5 times to ensure the stability of results. Furthermore, training was halted if the F1 score improvement over 200 mini-batches was less than the threshold $\delta = 1e-2$.

The convergence curves in Figure 2 illustrate the training speed of DAFOS. For all three datasets, DAFOS demonstrates faster convergence, reaching optimal performance in fewer epochs. This is particularly evident in the Reddit dataset, where DAFOS's curve is steeper, indicating it requires fewer epochs to achieve a comparable F1 score to SoTA-GCN. The rapid initial increase in F1 score for DAFOS can be attributed to both its node scoring mechanism, which prioritizes structurally important nodes early in training, and the dynamic fanout adjustment, which expands the model's receptive field as needed, leading to more efficient learning without unnecessary computation in the early stages. Overall, these curves highlight DAFOS's ability to converge faster while maintaining or improving accuracy compared to SoTA-GCN, showcasing the benefits of its adaptive sampling strategy.

The experimental results, as summarized in Table I, demonstrate that our proposed DAFOS-based GCN outperforms the SoTA-GCN model in both speed and accuracy across benchmark datasets. Our approach shows significant improvements in training efficiency, with the model converging faster while maintaining or improving the F1 score in most cases. On the ogbn-arxiv dataset, we achieve a 3.14x improvement in per-epoch speed and a 3.57x overall speed increase. For the Reddit dataset, our method provides an 8.5x faster per-epoch speed and a 12.6x overall improvement. Additionally, on the ogbn-products dataset, the per-epoch time and training time drop to 1299.99 ms and 744,355.1 ms, respectively. These substantial speed improvements are due to the dynamic fanout adjustment mechanism, which begins with a smaller fanout and expands it only when the model's loss plateaus, thereby minimizing excessive computation in the early stages and facilitating faster convergence while maintaining accuracy.

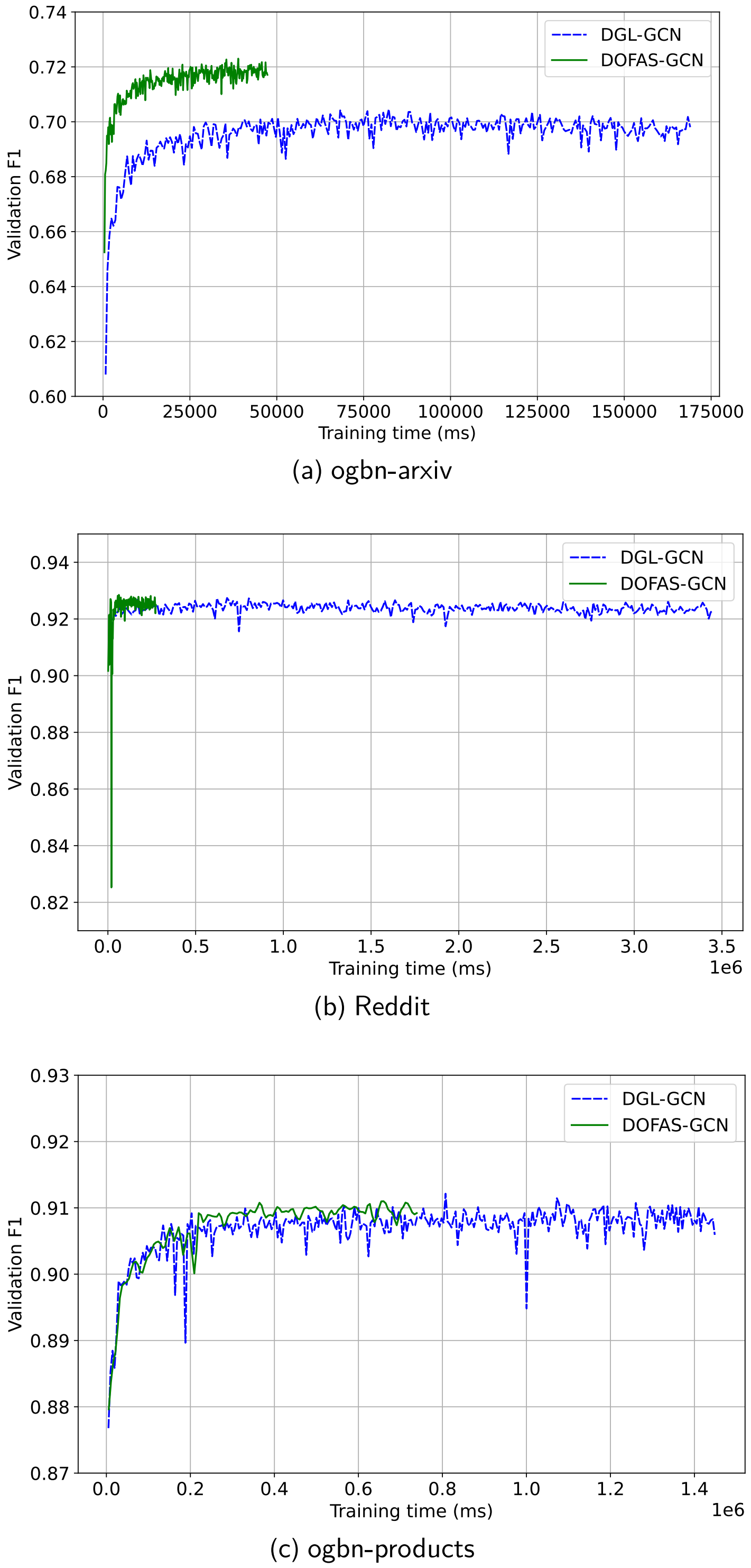

(a) ogbn-arxiv

(b) Reddit

(c) ogbn-products

Fig. 2: Convergence curves for validation F1-score and training time comparing SoTA and DAFOS GCNs on benchmark datasets.

[1]http://snap.stanford.edu/ogb/data/nodeproppred/arxiv.zip
[2]https://data.dgl.ai/dataset/reddit_self_loop.zip
[3]http://snap.stanford.edu/ogb/data/nodeproppred/products.zip
[4]https://github.com/dmlc/dgl/tree/master/examples

TABLE I: Comparison of SoTA and DAFOS GCNs on epoch time, total training time (in milliseconds (ms)), and F1 score (%).

| Models | Datasets | Epoch (ms) | Training (ms) | F1 score |
|---|---|---|---|---|
| SoTA-GCN | ogbn-arxiv | 563.27 | 168983 | 68.5 |
| | Reddit | 11491.15 | 3447345 | 92.1 |
| | ogbn-products | 3327.86 | 1448359.7 | 73.78 |
| DAFOS-GCN | ogbn-arxiv | 179.18 | 47304 | 71.21 |
| | Reddit | 1340.43 | 273449 | 90.3 |
| | ogbn-products | 1299.99 | 744355.1 | 76.88 |

TABLE II: Training time and F1 score for varying fanouts and loss thresholds.

| Fanout increment ($\Delta f$) | Loss Threshold ($\epsilon$) | Training Time (ms) | F1 score (%) |
|---|---|---|---|
| 5 | 0.0001 | 39111.27 | 71.7 |
| 5 | 0.001 | 42871.41 | 71.361 |
| 5 | 0.002 | 40584.78 | 71.781 |
| 5 | 0.005 | 51262.72 | 72.127 |
| 3 | 0.01 | 55265.69 | 72.043 |
| 5 | 0.01 | 47304 | 71.586 |
| 7 | 0.01 | 53401.09 | 71.731 |
| 9 | 0.01 | 53650.65 | 72.039 |

In addition to faster training times, our model also shows improved F1 scores on the ogbn-arxiv dataset, achieving 71.21% compared to 68.5%, and on the ogbn-products dataset, achieving 76.88% compared to 73.78%. DAFOS-GCN's node prioritization and dynamic fanout adjustment strategies prove especially effective in structured graphs like ogbn-arxiv and ogbn-products, resulting in higher F1 scores. The ogbn-arxiv dataset, being a citation network with a well-defined hierarchy, allows the model to leverage its node scoring mechanism by focusing on structurally important nodes early in training. This, combined with our dynamic fanout adjustment strategy, which gradually expands the model's receptive field as training progresses, ensures that influential nodes are prioritized and critical global information is captured more efficiently. By starting with a smaller fanout and expanding it based on model performance, the approach accelerates learning while optimizing computational resources, leading to improved generalization and higher overall accuracy.

In contrast, the Reddit dataset's homogeneous structure, where node connections are more uniform and high-degree nodes hold less importance, limits the effectiveness of DAFOS-GCN's dynamic strategies. Starting with a smaller fanout in the initial epochs further restricts the model's ability to gather global context early on, slowing its ability to represent the overall graph structure. By comparison, SoTA-GCN's fixed fanout consistently samples a broader set of neighbors throughout training, capturing more global information, which results in a slightly higher F1 score. However, despite this minor reduction in accuracy on Reddit, DAFOS-GCN significantly reduces training time, making it an efficient solution for large-scale datasets where balancing computational efficiency and accuracy is essential.

We also performed a sensitivity analysis on the ogbn-arxiv dataset, varying the $\Delta f$ and $\epsilon$ to assess their impact on training time and F1 score (Table II). Results show that increasing the $\Delta f$ leads to slightly higher F1 scores, with $\Delta f = 5$ providing a good balance between performance and efficiency, achieving an F1 score of 71.586% with a training time of 47,304 ms. Larger $\Delta f$ (e.g., nine) improves the F1 score marginally (72.039%) but increases training time. Similarly, stringent $\epsilon$ (e.g., 0.0001) results in faster training but slightly lower F1 scores, while higher $\epsilon$ (e.g., 0.01) leads to better performance at the cost of longer training times. Overall, a $\Delta f$ of five and an $\epsilon$ of 0.001 or 0.01 offer the best trade-off between speed and accuracy.

## V. Conclusion

In this paper, we presented the Dynamic Adaptive Fanout Optimization Sampler (DAFOS), an approach designed to enhance the efficiency and performance of GNNs by dynamically adjusting fanout based on model loss and prioritizing important nodes during training. By focusing on high-degree nodes early in the training process and expanding the fanout only when necessary, DAFOS accelerates model convergence while maintaining or improving accuracy. The experimental results on three benchmarks, i.e., ogbn-arxiv, Reddit, and ogbn-products demonstrate the effectiveness of the proposed approach, achieving significant reductions in both epoch time and total training time, with a substantial improvement in F1 score on ogbn-arxiv and ogbn-products datasets. The integration of dynamic fanout adjustment and early stopping ensures that DAFOS remains computationally efficient while effectively capturing both local and global graph structures. Our work highlights the potential of adaptive strategies in GNN training and opens new avenues for further optimization in large-scale graph learning.

In the future, we will extend our approach to test its effectiveness on other GNN models, such as GAT and GraphSAGE, applying it to other larger-scale datasets like ogbn-papers100M. This will allow us to evaluate the scalability and generalizability of DAFOS across different graph neural architectures and massive datasets, further refining its ability to handle diverse graph structures efficiently. Additionally, we aim to explore the integration of more complex node scoring mechanisms and adaptive sampling strategies to enhance performance on highly heterogeneous graphs.

## VI. Acknowledgements

This work was supported by Institute of Information & communications Technology Planning & Evaluation (IITP) grant funded by the Korea government(MSIT) (No.RS-2022-00155911, Artificial Intelligence Convergence Innovation Human Resources Development (Kyung Hee University)).